# ComplianceGate: Classifier-Gated Multi-Tier LLM Routing for Inference in Regulated Industries

**Abhishek Dey**
Independent AI Researcher, India
ORCID: 0009-0006-5427-7058
abicsen.info@gmail.com

## Abstract

Large language models deployed in regulated industries operate under two constraints: compliance enforcement and cost efficiency. Personally identifiable information (PII) in user queries can reach model endpoints before the system determines whether that data should leave its jurisdictional boundary. Serving all queries through a single large model consumes full GPU capacity regardless of query complexity while offering no mechanism for geographic routing. Mixture-of-Experts architectures do not address this routing occurs between expert layers within the model after data has already arrived at the endpoint, with all experts loaded in memory regardless of query complexity. We propose a classifier-gated routing architecture that enforces compliance by design. A trained encoder classifier sits before any decoder inference, evaluating each query for complexity and data sensitivity, then routing it to an appropriately sized dense model in the appropriate geographic location. PII-containing queries route to local endpoints before any LLM computation begins, eliminating data residency violations by design. Simple queries reach small, fast models at a fraction of the cost. Our evaluation of 600 queries demonstrates 39% mean latency reduction, 33-52% cost savings depending on query distribution, and generation throughput of 122-200 tokens/second versus 50-64 for the baseline. The encoder classifier achieves 99.2% accuracy with near-perfect PII recall at 7ms inference overhead, establishing pre-inference classification as a practical path to compliance-by-design LLM deployment.

## 1. Introduction

Large language models are increasingly adopted in financial services or regulated industries. However, deploying these models in production introduces a tradeoff between capability, cost, and regulatory compliance that current serving architectures do not address. Query distributions in financial services are heavily skewed, with the majority being straightforward lookups that do not require frontier-class models. Routing all queries to a single large model consumes full GPU capacity regardless of query complexity and creates infrastructure contention during peak demand. Regulated industries operate under data residency mandates: GDPR, SEBI, RBI, and India's DPDPA require that personally identifiable information must not cross jurisdictional boundaries during processing. Single-model deployments cannot enforce this. Mixture-of-Experts architectures do not address this at the serving layer - MoE routing occurs between expert layers within a single model at full GPU load, leaving no decision point for geographic enforcement once a query enters the model. Prior routing approaches address cost from different angles: FrugalGPT (Chen et al., 2023) cascades call through increasingly expensive models. RouteLLM (Ong et al., 2024) selects between two models based on quality preference. Hybrid LLM (Ding et al., 2024) routes between local and cloud endpoints. Each contributes to cost-efficient serving, but none enforce compliance per query - PII is not detected and data never routes based on sensitivity. We propose a classifier-gated routing architecture where a trained encoder classifier sits before decoder inference. The classifier, pre-trained on 2 billion tokens and fine-tuned on 50,000 domain-specific examples, classifies each query by complexity and data sensitivity, outputting a probability distribution with a confidence score that determines routing certainty. Simple non-sensitive queries flow to smaller models on cross-region infrastructure while PII-containing queries route to local endpoints before any LLM computation begins, enforcing compliance architecturally rather than through policy. This architecture simultaneously addresses:

**Compliance and Security:** PII is detected and contained before inference begins, ensuring sensitive data never leaves its jurisdictional boundary regardless of which model serves the response.

**Performance:** Simple queries route to smaller models with lower latency, eliminating the overhead of processing straightforward requests through frontier-class models.

**Cost Efficiency:** Only queries requiring complex reasoning consume expensive GPU capacity, reducing serving costs proportional to the actual complexity distribution of incoming traffic.

**Resiliency:** Each model tier operates independently with no shared failure domain. In case of inference failure or performance degradation in any tier, queries automatically fall back to other available models while maintaining compliance by design, ensuring continuous service without violating data residency constraints.

# 2. Architecture

## 2.1 System Overview

The system consists of two components operating within a private network: an encoder classifier and a routing layer. The encoder classifier runs on a GPU-equipped instance for sub-10ms inference. The routing layer runs on a CPU instance, performing classification interpretation, confidence-based fallback, and model tier selection with no model computation of its own. The request flow is a user query enters the routing layer, which calls the encoder classifier. The classifier returns a label and confidence score. The routing layer applies a confidence threshold, selects the target dense decoder model and geographic location, then forwards the original query for inference. The entire routing decision completes before any LLM computation begins. (Appendix-1)

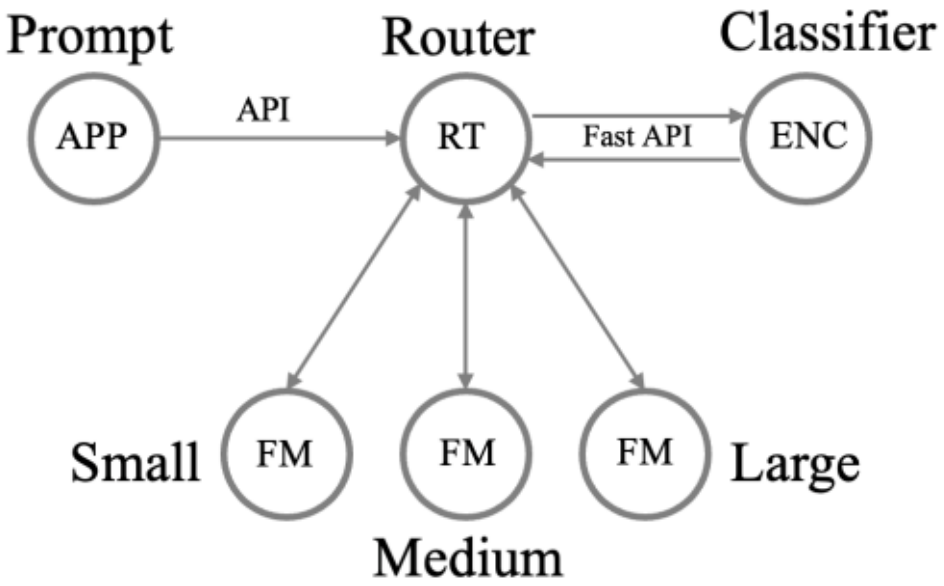

Figure 1: System Architecture – Classifier-Gated Routing

## 2.2 Encoder Classifier

The compliance classifier router is a bidirectional transformer encoder (Devlin et al., 2019) pre-trained with masked language modeling on 2 billion tokens of English text (500K training steps), then fine-tuned on 50,000 domain-specific examples for 6-class classification (12,000 steps). The architecture follows the standard encoder configuration: 12 layers, 768 hidden dimensions, 12 attention heads, and 3072 FFN dimensions, totaling approximately 134 million parameters. Input sequences are truncated at 128 tokens, sufficient for query classification where the full prompt is rarely longer.

The classifier takes raw query text as input and outputs a probability distribution over six labels. The classification head is a two-layer feed-forward network operating on the first token representation, with a tanh (hyperbolic tangent) activation between layers. Dropout is set to zero during inference. For this experiment, we pre-trained and fine-tuned the encoder from scratch; however, the architecture is agnostic to the classifier's origin. Practitioners may substitute any encoder configuration, use a publicly available pre-trained model, or adapt the fine-tuning dataset to their domain without modifying the routing framework.

## 2.3 Classification Labels and Routing Logic

The classifier outputs one of six labels representing the cross-product of three complexity levels and two sensitivity dimensions: The sensitivity dimension determines geographic routing queries containing PII are restricted to local infrastructure, enforcing data residency at the architectural level. The complexity dimension determines model tier: simple queries route to smaller, faster dense decoders, medium complexity queries route to mid-sized models balancing capability and cost, while complex queries route to larger, more capable ones.

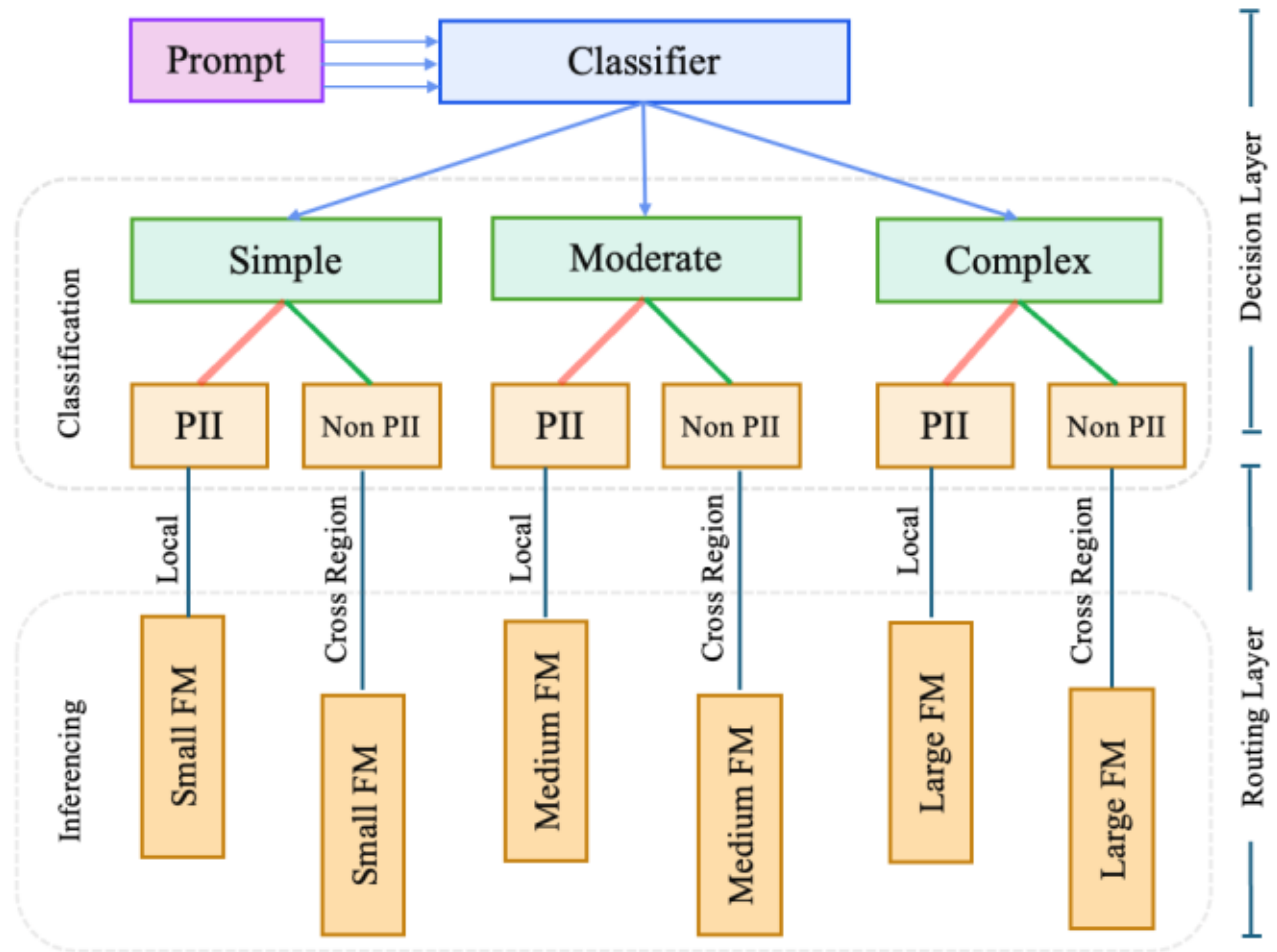

Figure 2: Classification Labels and Routing Logic

## 2.4 Confidence-Based Fallback

The classifier outputs a probability distribution over all labels; the highest probability value represents the model's confidence in its classification decision. When this confidence falls below a configurable threshold (default 80%), the system defaults to the most restrictive label (complex_pii). This guarantees that under uncertainty, queries receive full model capacity on local infrastructure. The fallback eliminates the risk of misrouting sensitive data to cross-region endpoints due to classifier uncertainty.

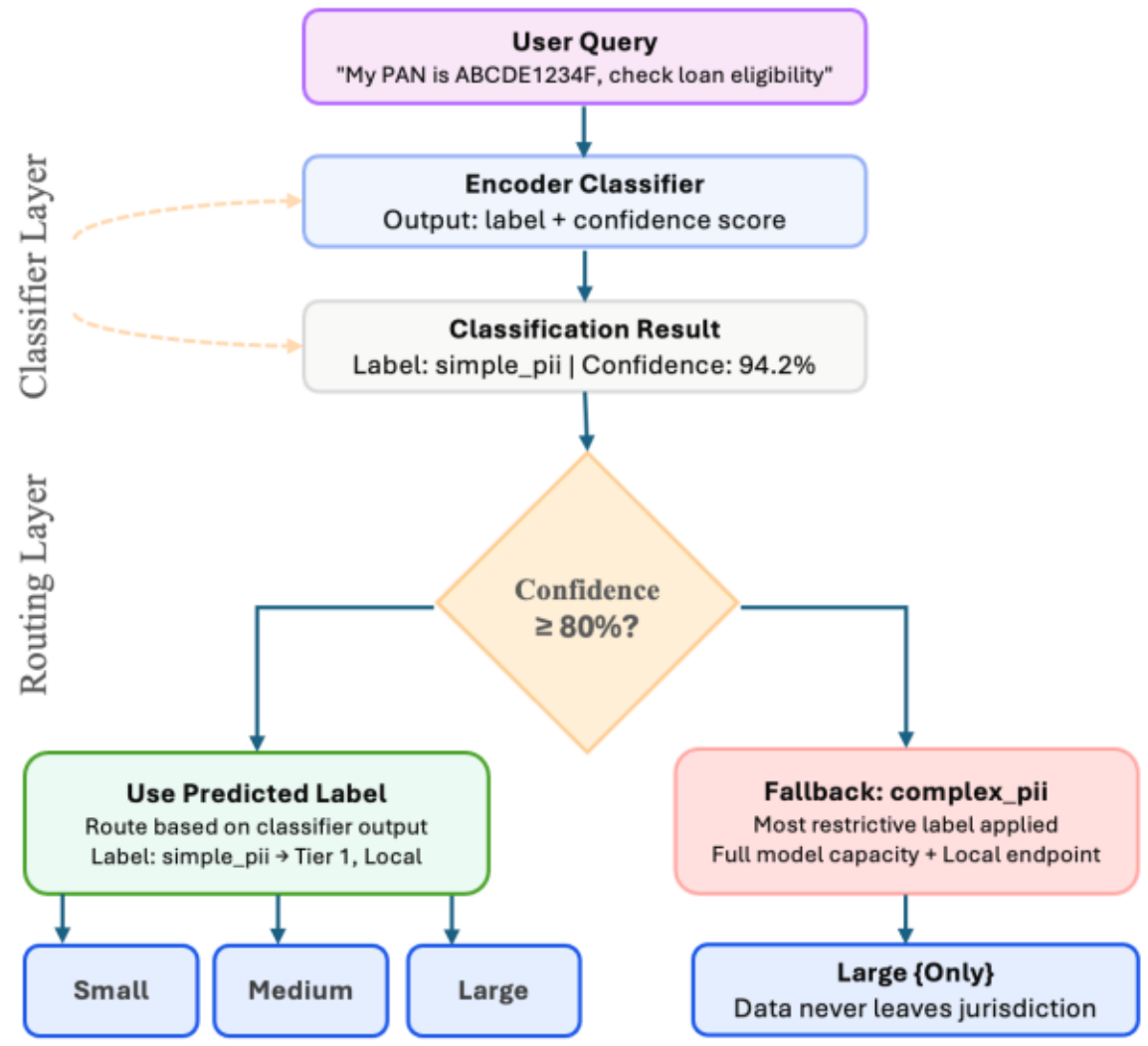

Figure 3: Confidence-Based Fallback Flow

## 2.5 Model Tiers

The architecture supports any number of dense decoder models behind the routing layer. Each tier operates independently on separate infrastructure with no shared memory or failure domain. Models can be of any size, provider, or architecture. The router holds endpoint references as externalized configuration, models can be swapped, upgraded, or scaled per tier without affecting other tiers or rebuilding the classifier.

### 2.6 Deployment

The encoder classifier and routing layer are packaged[1] as container images and deployed through infrastructure-as-code templates, enabling single-command deployment to any geographic location. The classifier container includes the model weights, tokenizer, and serving framework. The routing layer manages classification requests, confidence evaluation, and model tier selection through APIs. The full system can be replicated across jurisdictions.

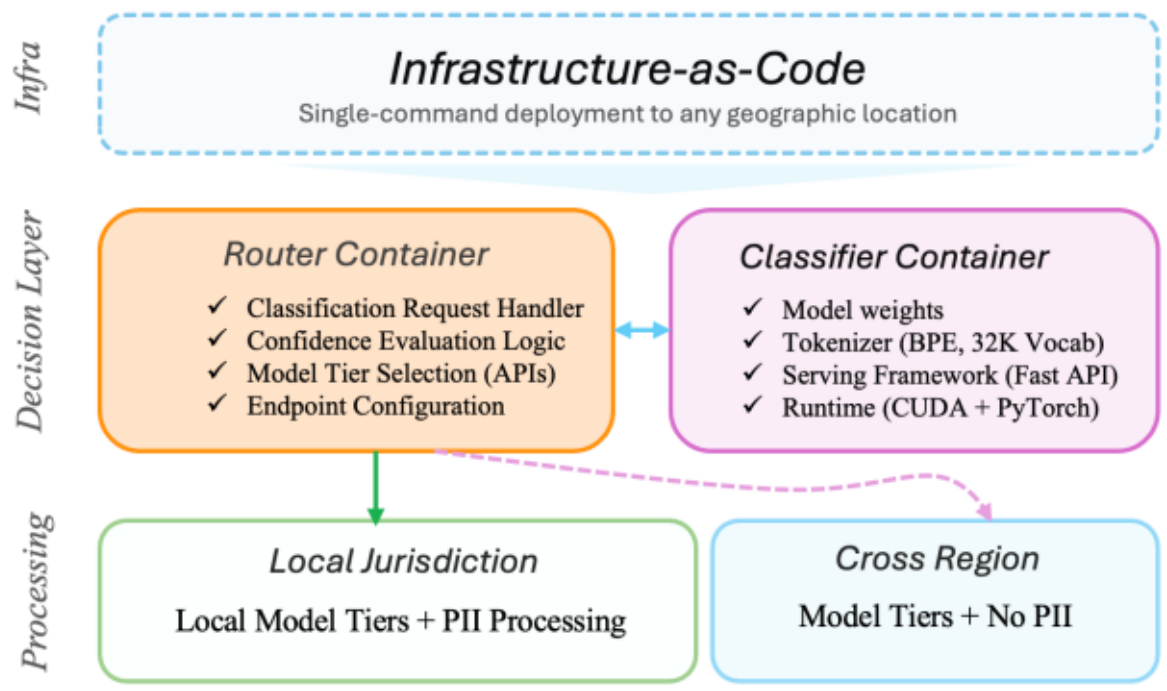

Figure 4: Deployment Architecture

## 3. Related Work

Several approaches have been proposed to reduce LLM serving costs through query routing, model cascading, or output mixing. We review the most relevant work and identify where each fall short for regulated industry deployment.

**FrugalGPT (Chen et al., 2023)** sends queries through a sequence of models from cheapest to most expensive, escalating only when a quality threshold is not met. This results in redundant computation on queries that require escalation and does not include PII detection or data residency controls.

**RouteLLM (Ong et al., 2024)** uses human preference data to train a router that selects between a stronger and a weaker model at inference time. It reduces cost effectively but is limited to two model tiers with no sensitivity classification or geographic routing.

**AutoMix (Madaan et al., 2024)** produces responses from both a small and a large model, then selects or combines the outputs after generation. This requires running both models for every query and includes no pre-inference content filtering or compliance mechanism.

**Hybrid LLM (Ding et al., 2024)** divides queries between a locally hosted small model and a cloud-hosted large model based on estimated difficulty. It supports only two routing targets and does not detect PII or enforce data residency.

**Mixture-of-Experts (Shazeer et al., 2017; Fedus et al., 2022; Jiang et al., 2024)** activates a subset of expert sub-networks per token within a single model. All experts share the same physical infrastructure, and routing decisions are made at the token level inside the model, which cannot inform geographic or model-level serving decisions.

Our work differs from all the above in combining trained classification with compliance enforcement in a single pre-inference decision. No existing approach simultaneously provides: (1) multi-tier model selection beyond binary routing, (2) PII detection integrated into the routing decision, (3) geographic enforcement based on content sensitivity, and (4) a formal cost model with measurable savings bounds. The closest work, RouteLLM, addresses cost through trained routing but operates without any compliance dimension.

## 4. Cost Model

We formalize the cost advantage of classifier-gated routing over single-model deployment. Let N denote the total number of queries over a given period, K the number of model tiers, and C_large represent the per-query inference cost of the most capable model in the system.

In a single-model deployment, all queries are served by the large model regardless of complexity:

$$C_single = N \times C_large$$

In our multi-tier architecture, queries are distributed across K tiers based on classification. Let $p_i$ denote the fraction of queries routed to tier i, where $\Sigma_{i=1}^{K} p_i = 1$, and let $C_i$ denote the per-query cost of tier i:

$$C_multi = C_classifier + N \times \Sigma_{i=1}^{K} (p_i \times C_i)$$

The classifier cost C_classifier is negligible: at under 7ms inference on a shared GPU instance serving 130 queries per second, the amortized per-query routing cost is below $0.00001, representing less than 0.1% overhead at any practical volume. We therefore simplify:

$$C_multi \approx N \times \Sigma_{i=1}^{K} (p_i \times C_i)$$

The fractional savings S relative to single-model deployment is:

$$S = 1 - (C_multi / C_single) = 1 - \Sigma_{i=1}^{K} (p_i \times C_i / C_large)$$

Let $r_i = C_i / C_large$ represent the cost ratio of tier i relative to the large model. Then:

$$S = 1 - \Sigma_{i=1}^{K} (p_i \times r_i)$$

We derive cost ratios from on-demand Bedrock pricing observed during our benchmark. Averaging across two models per tier: $r_1 = 0.15$ (simple tier, averaged from Nova Lite and Gemma 4B), $r_2 = 0.26$ (medium tier, averaged from Gemma 27B and Ministral 14B), $r_3 = 1.0$ (complex tier, same model as baseline).

For our 3-tier system (K=3) under equal benchmark distribution:

Distribution A ($p_1$=0.33, $p_2$=0.33, $p_3$=0.33) - experimentally validated:

$$S = 1 - (0.33 \times 0.15 + 0.33 \times 0.26 + 0.33 \times 1.0) = 1 - 0.466 = 0.52$$ (52% savings)

Under typical FSI query distributions where most requests are simple lookups:

Distribution B ($p_1$=0.60, $p_2$=0.25, $p_3$=0.15):

S = 1 - (0.60 × 0.15 + 0.25 × 0.26 + 0.15 × 1.0) = 1 - 0.305 = 0.695 (69.5% savings)

Distribution C ($p_1$=0.70, $p_2$=0.20, $p_3$=0.10):

S = 1 - (0.70 × 0.15 + 0.20 × 0.26 + 0.10 × 1.0) = 1 - 0.257 = 0.743 (74% savings)

The savings are bounded: as p_K approaches 1 (all queries are complex), savings approach zero. As $p_1$ approaches 1 (all queries are simple), savings approach (1 - $r_1$). The architecture provides maximum benefit when the query distribution is skewed toward simpler requests, which empirical data from financial services deployments consistently confirms.

At a scale of 1 million queries per month with C_large = $0.000727 per query (Qwen3-480B at 400 output tokens):

Single model: 1,000,000 × $0.000727 = $727/month
Multi-tier (Distribution A, equal): $727 × 0.466 = $339/month - savings $388 (52%)
Multi-tier (Distribution C, FSI): $727 × 0.257 = $187/month - savings $540 (74%)

The cost model assumes correct classification. In practice, misclassification affects savings in two directions: a complex query misrouted to a small model produces a low-quality response (quality cost, not monetary). A simple query misrouted to a large model wastes compute but produces a correct response. At 99.2% accuracy, misclassification affects less than 1% of queries, with negligible impact on aggregate cost savings.

# 5. Discussion

### 5.1 When Classifier-Gated Routing Works Best

This architecture provides cost and compliance benefits when: (a) the query distribution is skewed, with the majority being simple requests, (b) data residency enforcement is required per query, (c) each tier can scale independently, and (d) different model providers or sizes serve different tiers. The routing overhead is justified when C_classifier / C_model << 1. In our system, this ratio is 7ms / 500ms = 0.014 for the medium tier and 7ms / 2000ms = 0.0035 for the complex tier, confirming negligible overhead. These conditions hold in financial services, healthcare, and legal deployments.

### 5.2 When MoE is the Better Choice

MoE is preferable when queries do not separate into clean complexity categories, when a single model must handle all query types at the same quality level, or when the goal is training efficiency rather than serving cost reduction. MoE also handles ambiguous queries better since multiple experts can activate for a single token.

### 5.3 Sensitivity to Query Distribution

Cost savings are directly proportional to query distribution skew: shifting 10% of traffic from simple to complex reduces savings by approximately 8.5 percentage points (given $r_1$=0.15, $r_3$=1.0). The architecture provides diminishing returns as complex query proportion increases and reaches zero benefit when all queries require the large model.

### 5.4 Classifier Accuracy and Compliance Risk

A misclassified PII query (predicted as non-PII) could result in a data residency violation. The confidence-based fallback at 80% threshold mitigates this by routing uncertain predictions to the most restrictive tier. At 99.2% accuracy with near-perfect PII recall, fewer than 1% of queries are affected. For environments requiring zero tolerance, the threshold can be raised at the cost of routing more queries to the expensive local tier.

### 5.5 Future Work

Planned extensions include expanding to more than six classification labels for finer routing, multi-language PI detection beyond English, adaptive confidence thresholds based on observed error rates, distilled classifiers under 10M parameters for CPU-only deployment, and post-inference quality verification to re-route when a smaller model produces an insufficient response.

# 6. Comparison

The comparison uses three sources: GPU memory from published MoE specifications (Qwen et al., 2025), latency from our benchmark measurements, and cost from on-demand Amazon Bedrock pricing. Data residency, failure isolation, and scaling are provable from system design without benchmarks.

| Dimension | MoE | Classifier-Gated |
|---|---|---|
| Routing granularity | Per token, per layer | Per query, once before inference |
| GPU memory (simple queries) | 480 GB FP8 (all experts loaded) | ~4 GB (small model only) |
| Latency (simple queries) | 1,733–1,965ms | 490–840ms |
| Latency (complex queries) | 6,174–7,380ms | 6,845–7,347ms (same + 7ms overhead) |
| Routing overhead | 0ms (built into model) | 7ms per query |
| Per-query data residency | Not possible | Built-in (99.2% accuracy, ~100% PII recall) |
| Cost per simple query | $0.000180 | $0.000024–$0.000030 |
| Cost per complex query | $0.000727 | $0.000727 (same large model) |
| Average cost (70/20/10 split) | $0.000727 | $0.000339 (53% savings) |
| Failure isolation | Single point of failure | Independent tiers |
| Scaling | Scale entire cluster | Scale per tier |
| Geographic routing | Not possible | Per-query enforcement |

The table shows that classifier-gated routing provides structural advantages that MoE cannot replicate. Once a query enters the MoE model, geographic routing is no longer possible and tiers cannot scale independently. With gated routing, simple queries are served by models costing 6–8× less while consuming only 4 GB of memory versus 480 GB for the full MoE delivering equivalent answers at a fraction of the cost.

# 7. Experiments and Results

## 7.1 Experimental Setup

We evaluate two serving configurations on Amazon Bedrock on-demand inference in us-east-2, each processing 600 queries at temperature 0.3, top-k 10, top-p 0.9, with 100% success and zero failures. The query set spans three complexity tiers with controlled output lengths: 200 simple queries (100 tokens; banking FAQs, regulatory lookups), 200 medium queries (200 tokens; comparative analysis), and 200 complex queries (400 tokens; architectural design). The baseline routes all queries to Qwen3-Coder-480B-A35B regardless of complexity. The gated configuration routes to right-sized dense models: Amazon Nova Lite and Google Gemma 3 4B for simple, Google Gemma 3 27B and Mistral Ministral 14B for medium, and the same Qwen3-480B for complex. Two models per tier demonstrate model-agnostic routing and capture cross-provider variance.

## 7.2. Latency Comparison (P50)

Both configurations processed 600 queries with zero failures. At P50, classifier-gated routing reduces simple query latency by 2-4× (490-840ms vs 1850ms) and medium query latency by 1.7-3.3× (1140-2250ms vs 3750ms). Complex queries show no difference as both route to the same model. The gains target simple and medium tiers, which represent the majority of FSI traffic.

| Tier | Baseline (MoE 480B) | Gated (Right Sized) | Comparison / Improvement |
|---|---|---|---|
| Simple | ~1850ms | ~490-840ms | 2-4× faster |
| Medium | ~3750ms | ~1140-2250ms | 1.7-3.3× faster |
| Complex | ~6780ms | ~6850-7350ms | Same (same model) |

Table 1: Latency Comparison

## 7.3. Generation Throughput (Tokens / Second)

The throughput differential is attributable to model scale. Dense decoders at 4-27B parameters achieve 122-200 tok/s. The 480B MoE model, despite sparse activation, operates at 50-64 tok/s due to the memory footprint required to serve all expert parameters. Gated routing avoids this overhead for most queries by directing them to compact models.

| Tier | Baseline (MoE 480B) | Gated (Right Sized) |
|---|---|---|
| Simple | 50-55 tok/s | 122-200 tok/s (2-4× faster) |
| Medium | 56-59 tok/s | 87-175 tok/s (1.5-3× faster) |
| Complex | 64-69 tok/s | 62-64 tok/s (same model) |

Table 2: Inference Throughput Comparison Table

## 7.4. Latency Variance (Predictability)

The Max/Min ratio measures the spread between the fastest and slowest query within the same tier and model (derived from Table A1, Appendix-2). The MoE baseline exhibits ratios of 13–34×: a simple query completing in 635ms under favorable conditions can take over 21 seconds under contention - a 34-fold variance on identical inputs. Gated dense models show ratios of 1.1–2.0×, where the slowest query is at most twice the fastest. For systems bound by SLA contracts, a 34× variance cannot guarantee sub-second response times; a 1.7× variance can.

| Model | Tier | Max/Min Ratio | Interpretation |
|---|---|---|---|
| Qwen3 480B | Simple | 34.5x | Highly unpredictable |
| Qwen3 480B | Medium | 20.8x | Unpredictable |
| Qwen3 480B | Complex | 13.4x | Unpredictable |
| Nova Lite | Simple | 2.0x | Highly consistent |
| Gemma 4B | Simple | 1.7x | Highly consistent |
| Gemma 27B | Medium | 1.2x | Near constant |
| Ministral14B | Medium | 1.1x | Near constant |

Table 3: Latency Variance Max/Min

The P95/P50 ratio measures how predictable a model's response time is. A ratio close to 1.0 means nearly every query completes at the same speed. A high ratio means some queries take significantly longer than others under identical conditions. The MoE (Large) model shows P95/P50 ratios of 2.7-8.4×, indicating 5% of queries take 3-8 times longer than the median due to shared infrastructure contention. Gated dense models show 1.04-1.2×, near-constant performance. For FSI systems operating under SLA contracts, this predictability determines whether response time guarantees can be met reliably. Gated routing converts response time from a function of infrastructure contention into a function of query complexity.

| Model | Tier | P95/P50 Ratio | Interpretation |
|---|---|---|---|
| Qwen3 480B | Simple | 7.3x | Highly unpredictable |
| Qwen3 480B | Medium | 5.1x | Unpredictable |
| Qwen3 480B | Complex | 2.9x | Moderate variance |
| Nova Lite | Simple | 1.2x | Highly consistent |
| Gemma 4B | Simple | 1.1x | Highly consistent |
| Gemma 27B | Medium | 1.1x | Highly consistent |
| Ministral 14B | Medium | 1.03x | Near-constant latency |
| Qwen3 480B | Complex | 3.1x | Moderate variance |

Table 4: Latency Comparison P95/P50

## 7.5. Latency Distribution

Figure 5: Cumulative Latency Distribution (600 queries). Each curve shows the percentage of queries that completed within a given time. The green curve (gated routing) rises faster, meaning more queries finish quickly. At the 50%-mark, gated routing completes in 2,100ms while the MoE baseline takes 4,300ms. The gap is largest for simple and medium queries, which make up the majority of FSI traffic.

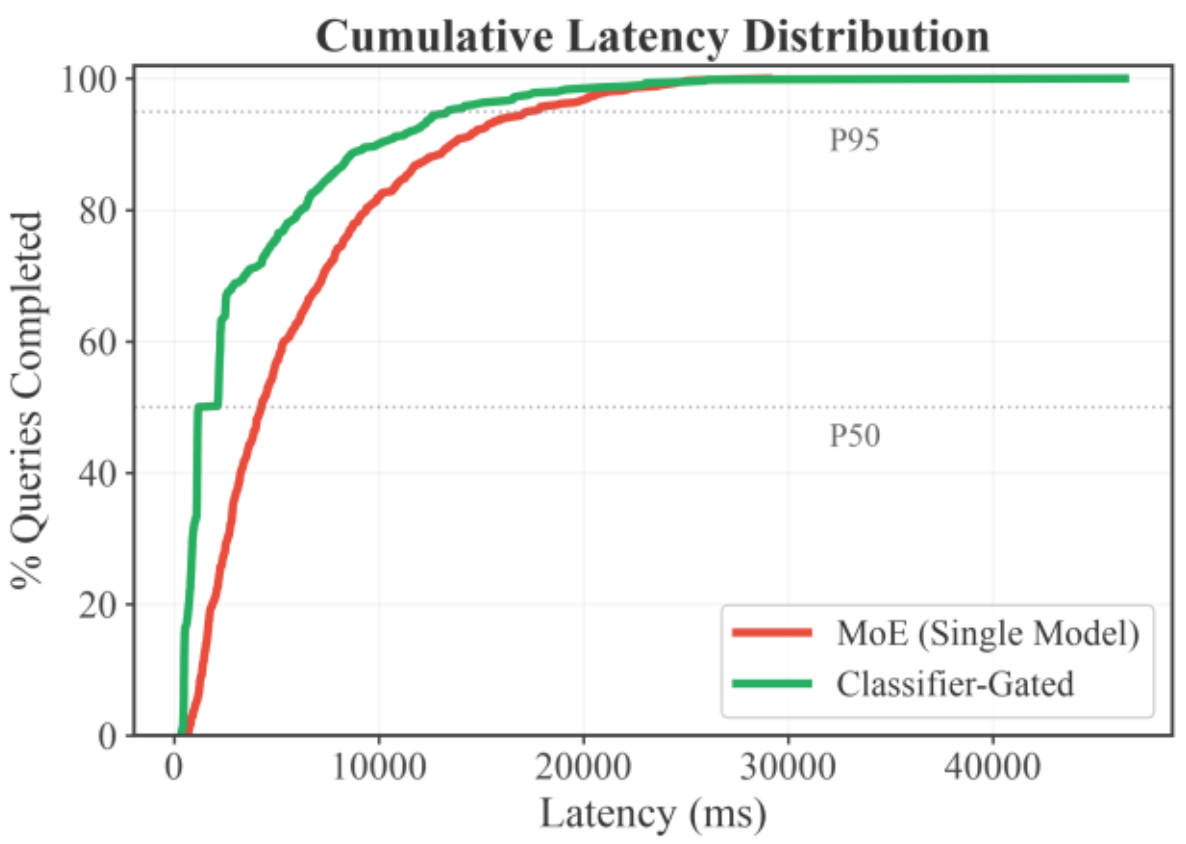

Figure 6: Per-Query Latency Sorted Ascending (600 queries). Each point represents one query. The MoE baseline (red) shows a long tail with queries exceeding 20,000ms, while gated routing (green) remains flat below 7,000ms for 80% of queries. The divergence beyond query 400 represents complex-tier traffic where both use the same model.

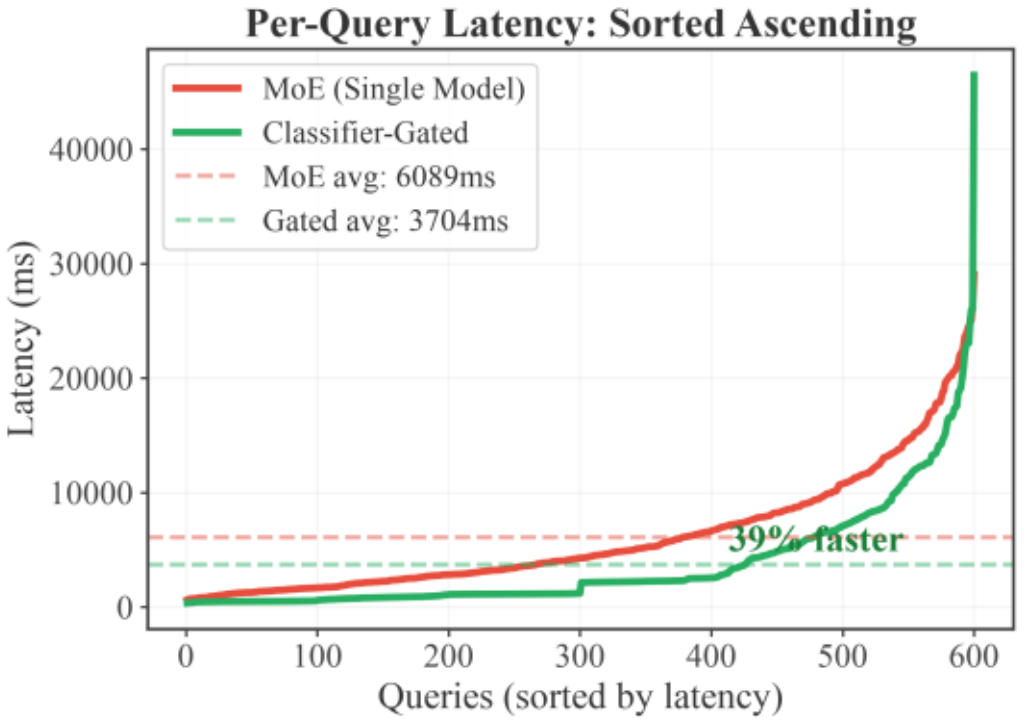

Figure 7: Cost per query under equal query distribution (33% simple, 33% medium, 33% complex). The shaded region represents cost avoided by routing simple and medium queries to smaller models. Under this balanced distribution, classifier-gated routing achieves 33% cost savings over the single-model baseline. Complex queries route to the same model in both configurations, producing identical responses.

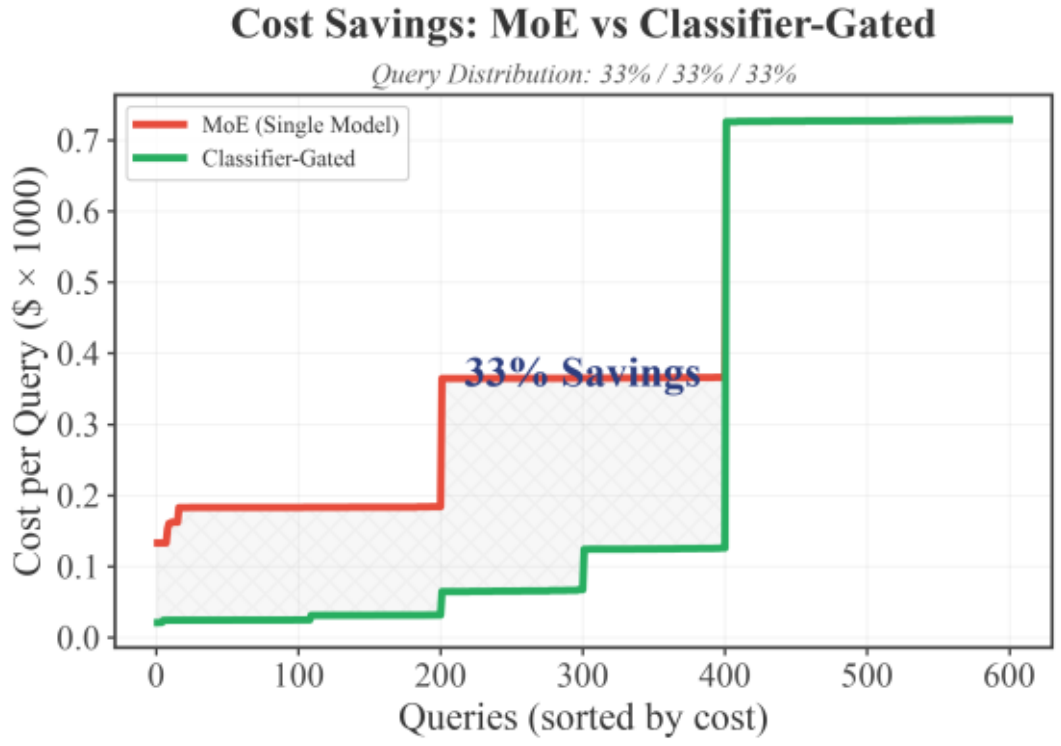

Figure 8: Cost per query under skewed query distribution (60/25/15), resampled from benchmark data. MoE requires all experts in memory (480 GB FP8) regardless of query simplicity, making per-query cost reduction impossible. The gated classifier routes simple queries to compact models (Nova Lite, Gemma 4B, Ministral 14B - 4 to 28 GB) at 6–8× lower cost per query. As the proportion of simple queries increases, savings expand to 52%.

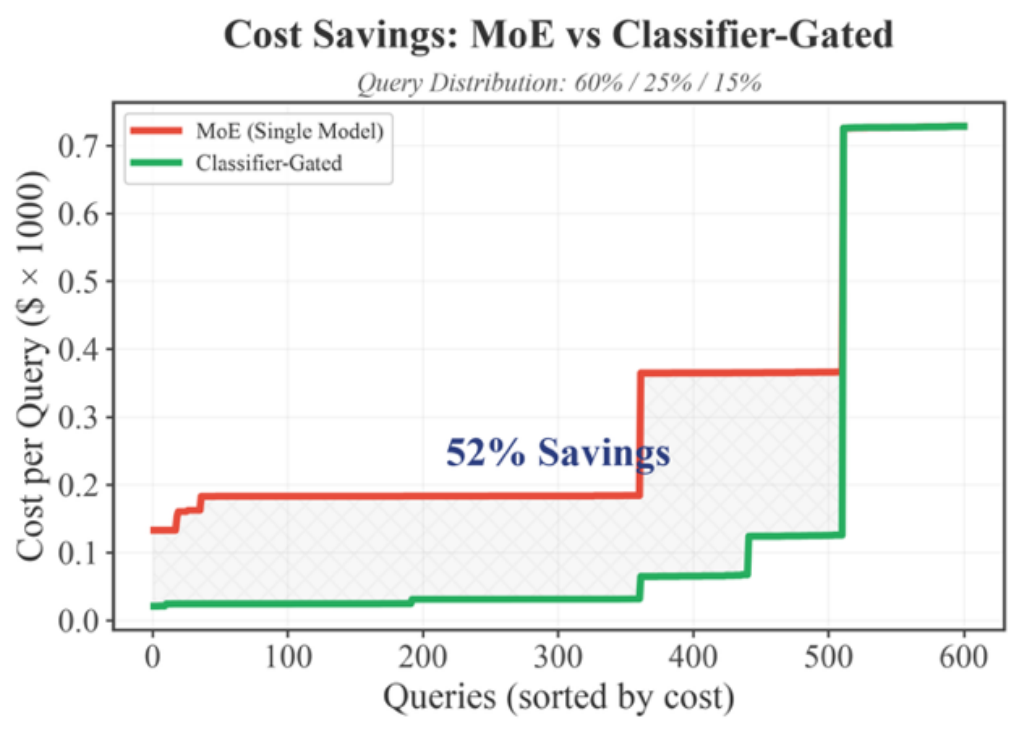

Figure 9: Latency Distribution by Tier. Each box shows the interquartile range (middle 50% of queries) with whiskers extending to the data range. MoE boxes are wide with long whiskers, indicating high variance, the same simple query can take anywhere from 635ms to 21,900ms. Gated boxes are compressed to near-flat lines, confirming that right-sized models deliver predictable latency regardless of load conditions.

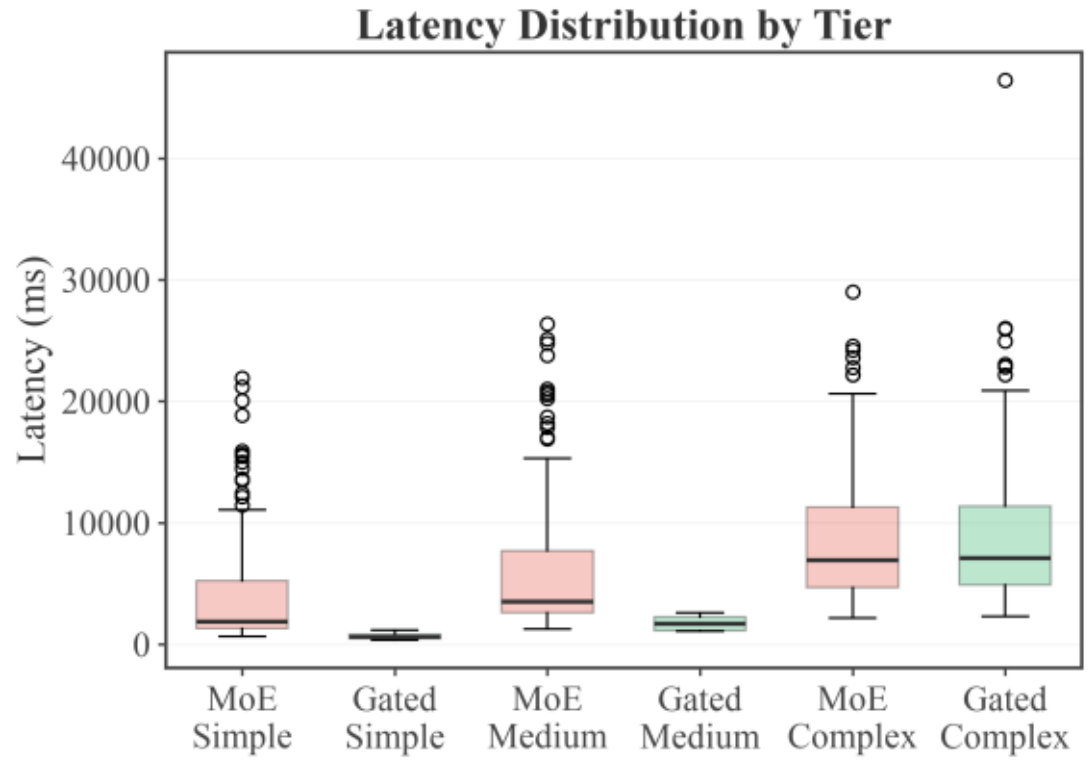

Figure 10: Time to First Byte (TTFB) CDF. TTFB measures how quickly the model begins generating output after receiving the query. Gated routing delivers first tokens faster for simple and medium queries because smaller models have less computational overhead before generation starts.

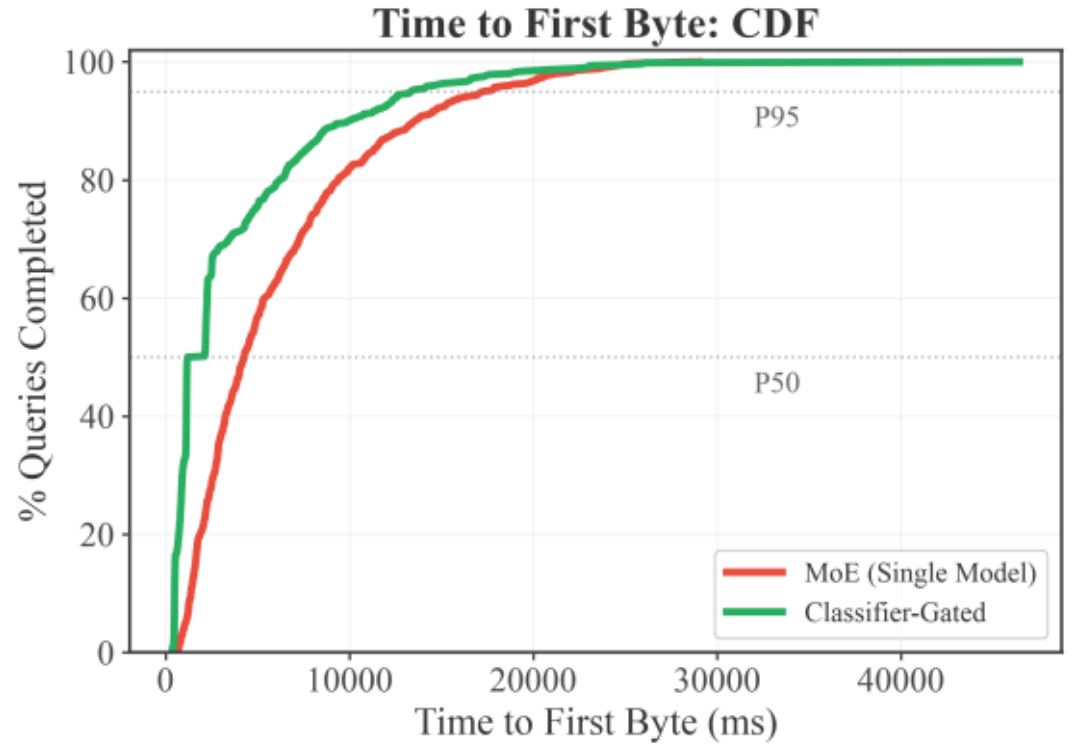

Figure 11: Generation Throughput (600 queries, sorted). Gated routing averages 119 tok/s versus 59 tok/s for MoE - 2× faster, driven by compact models on simple and medium tiers.

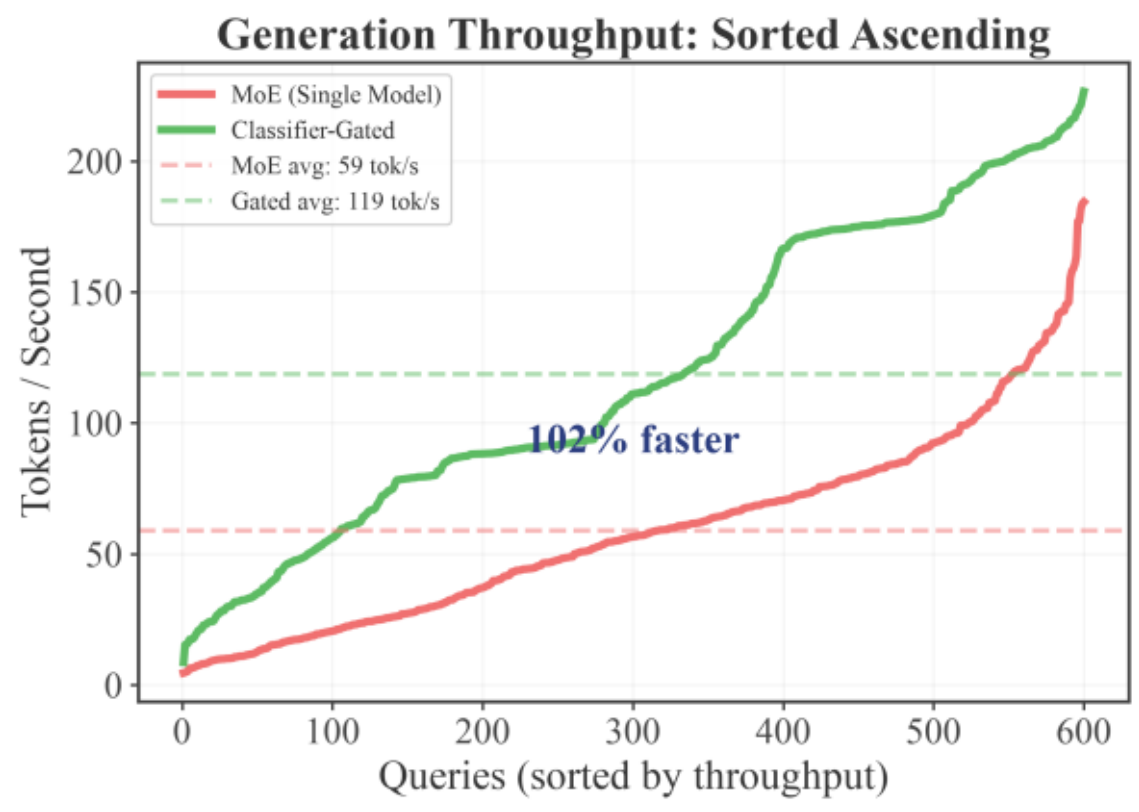

## 8. Conclusion

This work introduces a classifier-gated routing architecture that resolves compliance and cost at the pre-inference layer before query data reaches any model endpoint and before GPU resources are committed. Unlike Mixture-of-Experts, where all expert parameters must reside in memory and routing decisions occur internally after data arrives, our architecture evaluates each query before it enters any decoder and routes it to an appropriately sized dense model in the correct jurisdiction. Our benchmark on Amazon Bedrock (Qwen3-480B, Nova Lite, Gemma 4B, Gemma 27B, Ministral 14B, 600 queries, us-east-2) demonstrates 39% median latency reduction, with 2-4× improvement on simple queries and 1.7–3.3× on medium queries. Generation throughput reaches 122-200 tok/s versus 50-64 for the MoE baseline. Latency variance drops from 33× (Max/Min) to under 2×, making SLA guarantees feasible. Cost savings are 33% at equal query distribution and 52% under skewed distributions typical of regulated environments.

The encoder classifier achieves 99.2% accuracy on 6-class classification with near-perfect PII recall, trained from scratch on 2 billion tokens and fine-tuned on 50,000 domain-specific examples. Confidence-based fallback routes uncertain queries to the most restrictive tier, eliminating residency violations by design. The structural advantages per-query geographic enforcement, independent tier scaling, component-level upgrades, and full audit trails are properties that internal MoE routing cannot provide regardless of model quality. The system is containerized and deployable to any jurisdiction through infrastructure-as-code. By demonstrating that a trained encoder can enforce data residency, reduce serving costs by 33–52%, and maintain sub-10ms routing overhead with zero failures, this work provides a production-ready alternative to single-model and MoE deployments for regulated industries operating under strict compliance mandates.

This paper does not advocate for or against any specific model. The contribution is architectural: matching the right model to the right query based on measurable complexity and sensitivity. No single model is optimal for all queries, and no single deployment strategy satisfies both cost and compliance simultaneously. The value of this work lies not in which models are chosen, but in the routing framework that enables choosing them based on measurable complexity and sensitivity per query, per jurisdiction, per constraint. Future work targets finer classification labels, multi-language PII detection, adaptive confidence thresholds, and distilled classifiers under 10M parameters for CPU-only edge deployment.

## Appendix-1: System Architecture

The following diagrams illustrate the system architecture, request flow, classification logic, and deployment topology described in Section 2.

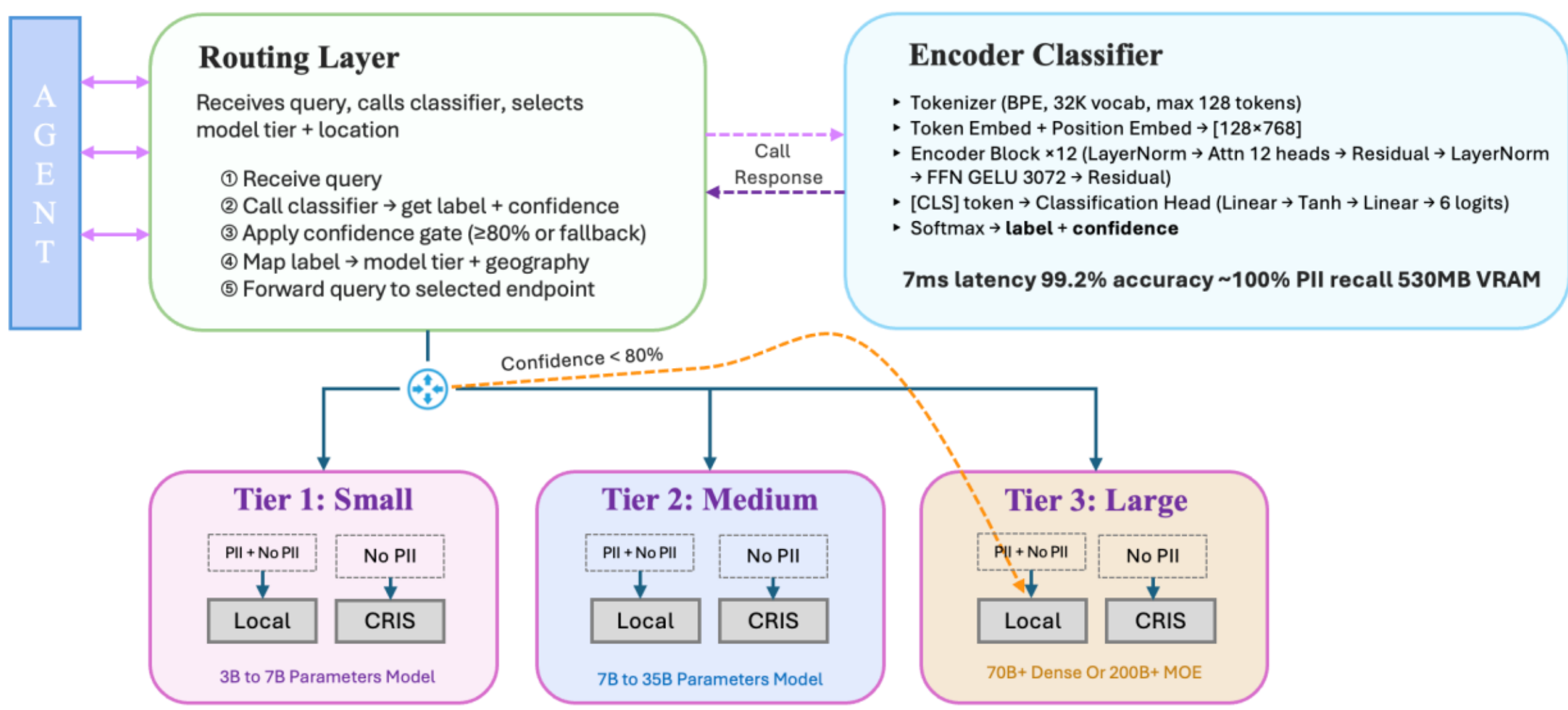

## Appendix-2: Detailed Benchmark Metrics

Table A1: Full Latency Distribution

| Model | Tier | P50 | P95 | P99 | Min | Max | Mean |
|---|---|---|---|---|---|---|---|
| Baseline – Qwen3 480B | Simple | 1733ms | 14506ms | 21205ms | 635ms | 21205ms | 3728ms |
| Baseline – Qwen3 480B | Medium | 3380ms | 15300ms | 23797ms | 1288ms | 23797ms | 5310ms |
| Baseline – Qwen3 480B | Complex | 7380ms | 19769ms | 29027ms | 2200ms | 29027ms | 8971ms |
| Gated – Nova Lite | Simple | 841ms | 1037ms | 1157ms | 582ms | 1157ms | 833ms |
| Gated – Gemma 4B | Simple | 492ms | 540ms | 608ms | 357ms | 608ms | 488ms |
| Gated – Gemma 27B | Medium | 2254ms | 2541ms | 2583ms | 2138ms | 2583ms | 2290ms |
| Gated – Ministral 14B | Medium | 1141ms | 1184ms | 1200ms | 1101ms | 1200ms | 1144ms |
| Gated - Qwen3 480B | Complex | 6845ms | 22174ms | 46445ms | 2618ms | 46445ms | 8837ms |

**Model** = inference model serving the query; **Tier** = query complexity level; **P50** = median latency (50th percentile); **P95** = 95th percentile latency (tail); **P99** = 99th percentile latency (worst-case tail); **Min** = fastest observed query; **Max** = slowest observed query; **Mean** = arithmetic average latency across all queries in that tier.

Table A2: Token Economics (Input/Output Efficiency)

| Config | Tier | Queries | Input Tokens | Output Tokens | Avg Input | Avg Output | Tok/s |
|---|---|---|---|---|---|---|---|
| Baseline – Qwen3 480B | Simple | 200 | 2,840 | 19,701 | 14.2 | 98.5 | 52.4 |
| Baseline – Qwen3 480B | Medium | 200 | 4,488 | 40,000 | 22.4 | 200.0 | 57.3 |
| Baseline – Qwen3 480B | Complex | 200 | 6,504 | 80,000 | 32.5 | 400.0 | 66.9 |
| Gated – Nova Lite | Simple | 100 | 600 | 10,000 | 6.0 | 100.0 | 112.3 |
| Gated – Gemma 4B | Simple | 100 | 1,496 | 9,766 | 15.0 | 97.7 | 200.3 |
| Gated – Gemma 27B | Medium | 100 | 2,312 | 20,000 | 23.1 | 200.0 | 87.6 |
| Gated – Ministral 14B | Medium | 100 | 1,760 | 20,000 | 17.6 | 200.0 | 174.9 |
| Gated - Qwen3 480B | Complex | 200 | 6,504 | 80,000 | 32.5 | 400.0 | 63.2 |

**Config** = serving configuration and model used; **Tier** = query complexity level; **Queries** = number of queries executed; **Input Tokens** = total prompt tokens across all queries in the tier; **Output Tokens** = total tokens generated across all queries; **Avg Input** = mean prompt length per query (tokens); **Avg Output** = mean generation length per query (tokens); **Tok/s** = average generation throughput per query (tokens per second).

---

[1] The pre-trained encoder classifier released at: https://huggingface.co/deycoding/deycoding-compliance-classifier-router.
Deployment code and infrastructure templates: https://github.com/deycoding/deycoding-compliance-classifier-router